\documentclass[11pt,a4paper]{article}

\usepackage[hyperref]{eacl2021}

\usepackage{times}
\usepackage{latexsym}
\usepackage[utf8]{inputenc}
\usepackage[T1]{fontenc}
\usepackage{booktabs}

\newcommand{\midsepremove}{\aboverulesep = 0mm \belowrulesep = 0mm}
\midsepremove
\newcommand{\midsepdefault}{\aboverulesep = 0.200mm \belowrulesep = 0.400mm}
\midsepdefault

\usepackage{amsfonts}
\usepackage{nicefrac}
\usepackage{microtype}
\usepackage[colorinlistoftodos]{todonotes}
\usepackage{float}
\usepackage{amsmath}
\usepackage{amssymb}
\usepackage[d]{esvect}
\usepackage{multirow}
\usepackage[normalem]{ulem}
\useunder{\uline}{\ul}{}
\usepackage{wasysym}
\usepackage{graphicx} 
\usepackage{mathtools}
\usepackage{caption}
\usepackage{mwe}
\usepackage{subfigure}
\usepackage{comment}
\usepackage{wrapfig}
\usepackage{tikz}
\usepackage{ctable}
\usepackage{array}
\newcolumntype{?}{!{\vrule width 0.1em}}
\usepackage{graphics}
\usepackage{adjustbox}
\usepackage{rotating}
\usepackage{graphicx}
\usepackage{url}

\usepackage{paralist}
\usepackage[capitalise]{cleveref}

\aclfinalcopy
\def\aclpaperid{884}

\makeatletter
\def\@fnsymbol#1{\ensuremath{\ifcase#1\or \dagger\or \ddagger\or
   \mathsection\or \mathparagraph\or \|\or **\or \dagger\dagger
   \or \ddagger\ddagger \else\@ctrerr\fi}}
\makeatother

\makeatletter
\newcommand{\printfnsymbol}[1]{%
  \textsuperscript{\@fnsymbol{#1}}%
}
\makeatother

\author{Daniel de Vassimon Manela\thanks{\ \ Equal contribution.} \qquad David Errington\printfnsymbol{1} \qquad Thomas Fisher\printfnsymbol{1} \\
\large\bf Boris van Breugel\printfnsymbol{1} \qquad Pasquale Minervini \\
  University College London \\
  \texttt{\{daniel.manela.19,david.errington.19,thomas.fisher.19,} \\
  \texttt{boris.breugel.19,p.minervini\}@ucl.ac.uk}
}

\title{Stereotype and Skew: Quantifying Gender Bias \\ in Pre-trained and Fine-tuned Language Models}

\date{}

\begin{document}

\maketitle

\begin{abstract}
This paper proposes two intuitive metrics, \textit{skew} and \textit{stereotype}, that quantify and analyse the gender bias present in contextual language models when tackling the WinoBias pronoun resolution task. We find evidence that gender stereotype correlates approximately negatively with gender skew in out-of-the-box models, suggesting that there is a trade-off between these two forms of bias. We investigate two methods to mitigate bias. The first approach is an online method which is effective at removing skew at the expense of stereotype. The second, inspired by previous work on ELMo, involves the fine-tuning of BERT using an augmented gender-balanced dataset. We show that this reduces both skew and stereotype relative to its unaugmented fine-tuned counterpart. However, we find that existing gender bias benchmarks do not fully probe professional bias as pronoun resolution may be obfuscated by cross-correlations from other manifestations of gender prejudice. Our code is available \href{https://github.com/12kleingordon34/NLP_masters_project}{online}.
\end{abstract}

\section{Introduction}\label{Section:intro}
Transformer-Based Transfer Learning models for NLP -- referred to henceforth as TBTL models for brevity -- such as 
BERT~\citep{bert}, RoBERTa~\citep{roberta}, and ALBERT~\citep{DBLP:conf/iclr/LanCGGSS20} perform well on a variety of NLP tasks with minimal fine-tuning.
However, prior to finetuning, TBTL models require a vast amount of data to train~\citep{megatron}.
This training is only performed once, with users downloading and fine-tuning such language models to their specific task.
In doing so, we are trusting large tech companies to train the base model responsibly since we have no control over this. This seems inherently undemocratic.
We ideally want these 
models to be free from unwanted bias and whilst it is true that they exhibit less gender bias than static word embeddings~\citep{sunlitreview}, they are by no means immune to this problem~\citep{Lu_contextual_bias}.
As TBTL models become increasingly prevalent in our everyday lives, we want to avoid such prejudices influencing decision making.
Examples where this is important include automatic resume filtering~\citep{dastin} and criminal sentencing recommendations~\citep{tashea_17AD}.
In this paper we focus on the specific problem of gender bias, and analyse the extent to which it persists in modern TBTL models.
We build upon \citet{zhao2018gender,zhao2019gender}, in which quantification and mitigation of bias in ELMo was centre stage. In addressing this problem for more recent models, we aim to answer three main questions:
\begin{inparaenum}[i)]
\item How can we quantify bias in pre-trained language models?
\item How do different models compare in terms of bias?
\item How to mitigate bias in these models?
\end{inparaenum}
We believe that current gender bias metrics in the existing literature do not offer sufficient granularity to properly analyse this problem.
Indeed, they mostly focus on measuring the assignment of stereotypical pronouns to professions \citep{zhao2018gender}. 
By focusing solely on this, they fail to address a model's overall preference for predicting male pronouns.
An alternative bias which models can demonstrate is unequal preference towards male and female pronoun resolution across stereotypical and anti-stereotypical professions.
We refer to these two forms of bias as \emph{skew} and \emph{stereotype}, respectively. In \cref{Sec:method}, we propose a new scheme to capture and quantify the important distinction between the two.
When comparing different TBTL models, we find evidence that gender skew and gender stereotype correlate approximately negatively with each other in out-of-the-box models, suggesting that a tradeoff between these two forms of bias may exist.

To mitigate bias in these models, we use the method proposed by \citet{zhao2018gender} to show that fine-tuning with augmented data, which references male and female entities with equal frequencies, can reduce professional gender stereotype and skew compared to fine-tuning on the original dataset.
However, we show that gender prejudice may persist in forms other than professional bias, and these are ineffectively probed by current NLP benchmarks.  %

\section{Related work}\label{sec:relatedwork}

\paragraph{Bias quantification}

Early work in measuring gender bias specifically \citep{Caliskan,may2019measuring}, along with efforts towards removing it either during~\citep{zhao2018glove} or after training~\citep{bolukbasi2016man}, was done on static word embeddings such as GloVe and Word2Vec. 

\citet{Caliskan} argue that completely removing undesirable bias using an automated procedure is impossible, as it is only distinguishable from the rules and structure of language itself by negative consequences in downstream applications.
Instead, we should focus on probing and exposing which biases manifest themselves in which models, so that engineers can act accordingly.
Choosing a suitable metric with which to analyse bias is a key challenge~\citep{may2019measuring}; whilst a positive result with respect to a suitable metric does reveal the existence of bias, a negative result does not mean a model is completely bias-free.

\begin{figure*}[t]
    \centering
    \includegraphics[width=\textwidth]{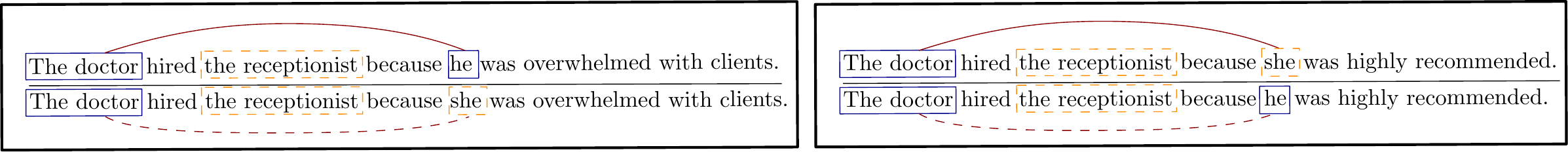}
    \caption{Example sentences from Test Set 1 (T1) of the WinoBias dataset. These take the form \textit{[entity1] [interacts with] [entity2] [conjunction] [pronoun] [circumstances]}. Solid blue boxes indicate male entities, dashed orange boxes indicate female entities, solid purple lines indicate pro-stereotypical scenarios and dashed purple lines indicate anti-stereotypical scenarios. Such stereotypes are determined according to \citet{laborstats}.
    }
    \label{fig:winobias1}
\end{figure*}

\begin{figure*}[t]
    \centering
    \includegraphics[width=\textwidth]{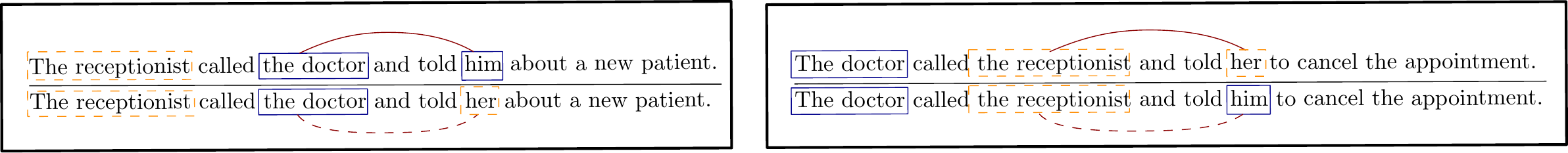}
    \caption{Example sentences from Test Set 2 (T2) of the WinoBias dataset. These take the form \textit{ [entity1] [interacts with] [entity2] and then [interacts with] [pronoun] for [circumstances]}. The style format follows that of \cref{fig:winobias1}. Image adapted from \citet{zhao2018gender}.}
    \label{fig:winobias2}
\end{figure*}

Statistical tests such as Word/Sentence Embedding Association Tests (WEAT/SEAT) have been developed to measure bias in static word embeddings using the cosine similarity of specific target words~\citep{Caliskan, may2019measuring}.
However, when it comes to contextual embeddings, these traditional metrics have been shown to be ineffective at quantifying bias.
In particular, \citet{kurita2019measuring} demonstrate that while WEAT tests are unable to identify any statistically significant bias in BERT, probing the underlying language model with a Gender Pronoun Resolution (GPR) task does reveal strong evidence that these non-static models also encode gender bias.
Indeed, since contextual word embedding models such as BERT are optimised to capture the statistical properties of training data, they tend to pick up and amplify any social stereotypes that may be present~\citep{kurita2019measuring}.
Having established GPR as a downstream task suitable for detecting gender bias, \citet{zhao2018gender} introduced a new benchmark, WinoBias, to measure bias in coreference pronoun resolution. 
The dataset consists of two files, Test Set 1 and Test Set 2 (hereafter T1 and T2), representing two different gender pronoun resolution tasks.
Each file consists of Wino-grad schema pairs of sentences involving a variety of occupations, differing only in one or two words and with a pronoun ambiguity that is resolved in opposite directions across the two sentences, giving both a pro- and anti-stereotypical resolution~\citep{levesque_winograd_2012}.
Example sentences are shown in \cref{fig:winobias1} and \cref{fig:winobias2}.
\citet{sunlitreview} consider a coreference resolution system unbiased on the WinoBias test if it achieves similar F1 scores for gender pronoun resolution on both the pro- and anti-stereotypical datasets whilst maintaining strong GPR performance.
One of the main findings in \citet{zhao2018gender} is that three different coreference resolution architectures (rule based, feature-rich and neural-net based) built on top of static word embeddings all display significant disparity in F1 scores across the two datasets, with the F1 score for the pro-stereotypical dataset being on average 21.1 higher.
This alarming observation was also discovered by \citet{webster2018mind} and was attributed to the inherent bias of the underlying word embeddings~\citep{bolukbasi2016man}, as well as the training of these coreference resolution pipelines on the OntoNotes 5.0 dataset~\citep{ontonotes} which is known to suffer from severe gender imbalance~\citep{zhao2018gender}.
More recently, \citet{zhao2019gender} investigated the existence of gender bias in the ELMo contextual embedding.
Specifically, they note ELMo is trained on the Billion Word corpus~\citep{1billionwords} which, just like OntoNotes 5.0, shows substantial imbalance in the counts of male vs. female pronouns.
Training on this, ELMo then learns a language representation that reflects this gender inequality.
To expose this, \citet{zhao2019gender} analyse the behaviour of a coreference resolution system proposed by \citet{kentonleecoref} with ELMo contextual weights on the WinoBias benchmark, revealing a significant disparity in performance on the pro- and anti-stereotypical datasets. In fact, this disparity is 30\% higher than a similar result based only on GloVe embeddings~\citep{leegloveresult}.
This is particularly worrying; as commented earlier, contextual embeddings may, by construction, be amplifying undesirable statistical artefacts of the dataset more than their static counterparts.
Therefore, it is of the utmost importance to perform a similar analysis on recent TBTL models.
\citet{bert} explicitly state that BERT is not trained on the Billion Word corpus, since this only provides examples of isolated sentences and the authors preferred to use a document-level corpus to get contiguous training data, allowing richer contexts to be learnt.

Specifically, BERT is trained using the BookCorpus dataset~\citep{bookcorpus} as well as English Wikipedia.
However, the BookCorpus data has since been shown to suffer from similar gender imbalance problems~\citep{tan2019assessing} as has English Wikipedia where, for example, only 15.5\% of the biographies are of women~\citep{Wagner_2016}. We believe that this imbalance is the principle cause of skew in the model.
\paragraph{Bias Mitigation}
As discussed above, whilst we cannot completely remove bias from a model, research into bias mitigation is still a very worthwhile pursuit and, in the context of the WinoBias metric of occupational gender bias, could help break the glass ceiling.
Many bias mitigation methods for static embeddings centre around modifying the vector space and/or loss function during the training process.
Initial attempts sought to project biased embedding vectors back to a gender neutral subspace~\citep{bolukbasi2016man}.
Subsequent improvements came from adding a regularisation term to the training loss function designed to encourage specific gendered words to separate, thus allowing the remaining neutral terms to mix~\citep{zhao2018glove}.
However, these offer superficial reductions in gender bias, and systematic prejudice was found to persist~\citep{gonen2019lipstick}.
Attempts to mitigate gender bias in contextualised embeddings are a more novel endeavour. These attempts typically involve fine-tuning models to a particular task and one proposal involves duplicating the training corpus and switching gender-specific terms in the duplicated data. For example, ``The King cemented his rule over his lords'' is substituted with ``The Queen cemented her rule over her ladies''. This method, referred to as Data Augmentation, was demonstrated to successfully reduce gender bias in ELMo for pronoun resolution tasks, relative to a model trained on the unaugmented training data \citep{zhao2019gender}.

\section{Method} \label{Sec:method}

\subsection{Analysing Bias in WinoBias} \label{subsec:method1}

\begin{figure}[t]
\begin{center}
\includegraphics[width=0.45\textwidth]{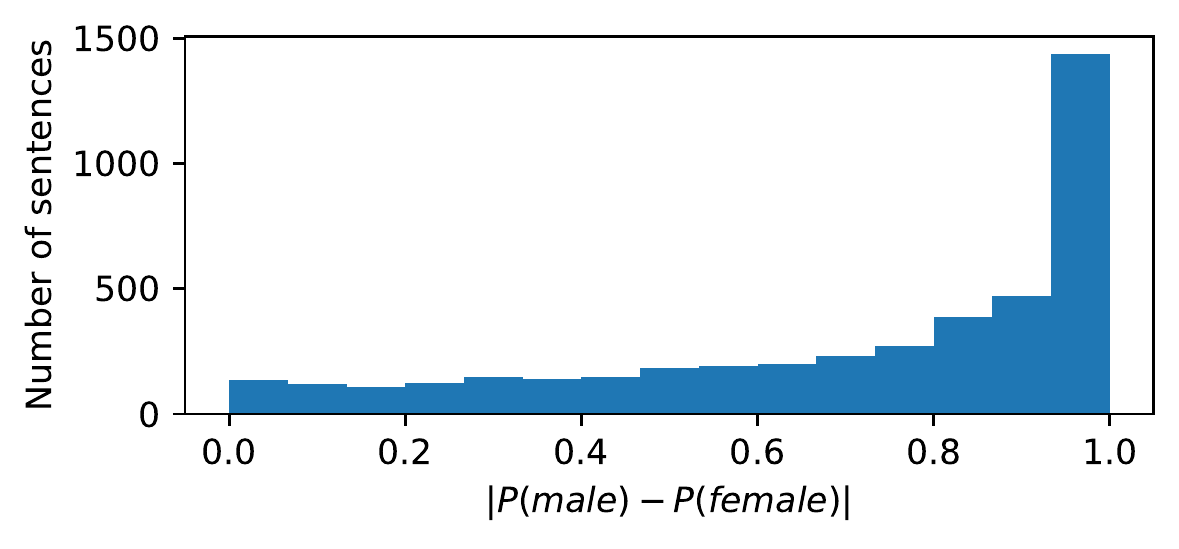}
\caption{Histogram of absolute differences in gendered pronoun assignment, $|P(\text{male})-P(\text{female})|$. Sentence examples where this difference was smaller than 0.1 were removed so that only pronouns assigned with a high degree of certainty were analysed.}
\label{fig:histogram}
\end{center}
\end{figure}

Bias in TBTL models can be measured using either T1 or T2 from the WinoBias dataset -- see \cref{fig:winobias1} and \cref{fig:winobias2}, respectively.
Within each test set, examples are composed of a pro-stereotypical and anti-stereotypical sentence, where stereotype is determined by professional gender imbalances recorded by \citet{laborstats}.
Our approach is to take each WinoBias sentence, mask the pronoun of interest, and then compare the language model's prediction for the masked token with the pro- and anti-stereotypical labels.
To predict the gender of the pronoun in the sentence ``\emph{The physician hired the secretary because [MASK] was overwhelmed with clients}'' we calculate the probabilities of the pro- and anti-stereotypical pronouns -- ``he'' and ``she'' respectively -- and pick the one with the highest likelihood. Note that this approach risks obscuring the confidence in pronoun resolution. For example, $P(\text{male}) = 0.99$ and $P(\text{male}) = 0.51$ would both result in male pronoun assignment. The histogram in \cref{fig:histogram} demonstrates that this issue does not affect our experiments; the distribution is highly negatively skewed. The majority of pronouns are resolved with a high degree of confidence. We chose $|P(\text{male})-P(\text{female})| \geq 0.1$ as an arbitrary cutoff bound to select sentences which were resolved with a high degree of certainty. Only sentences which fulfil this criteria were analysed in the following experiments.

In line with the academic literature, we compute F1 scores for both the pro- and anti-stereotypical data using contextual language models.
This approach to coreference resolution is demonstrably well-founded and F1 results from the GPR baseline are discussed in \cref{subsec:alicebob}.
The WinoBias sentences have been constructed so that, in the absence of professional stereotypes, there is no objective way to choose between different gender pronouns.
The difference in F1 scores with respect to gender $g$, across a pro/anti test set, $\text{F1}^g_\text{pro} - \text{F1}^g_\text{anti}$, is a metric inspired by previous papers to measure a model's tendency to assign that gender to professions, with positive (resp. negative) values indicating a pro- (resp. anti-) stereotypical assignment~\citep{sunlitreview}.
We refer to it as a measure of gender \emph{stereotype}. In contrast to the literature, we compute F1 scores with respect to both ``male'' and ``female'' true labels allowing us to define \emph{stereotype} with respect to both genders.
We now propose to also use the difference in F1 scores with respect to a dataset $\mathcal{D}$, across genders, $ \text{F1}^{\mars}_\mathcal{D} - \text{F1}^{\female}_\mathcal{D}$, as a measure of gender \emph{skew} in dataset $\mathcal{D}$, with positive (resp. negative) values capturing the tendency of a model to generally assign a male (resp. female) gender to any given profession.
This distinction is important: consider a classifier which only assigns male pronouns to professions. 
It would not be stereotyping professions to perceived gender roles, but would be heavily biased in assuming a general male dominance in the workplace.
Both these forms of gender unfairness are considered in the subsequent analysis and we use the mean skew and stereotype, taken across datasets and genders respectively as shown below:
\begin{align*}
     \mu_\text{Skew} \ \triangleq \ & \frac{1}{2} \left(\left|\text{F1}^{\mars}_\text{pro}  -\text{F1}^{\female}_\text{pro} \right| + \left| \text{F1}^{\mars}_\text{anti}  - \text{F1}^{\female}_\text{anti} \right| \right) \\
    \mu_\text{Stereo} \ \triangleq \ & \frac{1}{2} \left( \left| \text{F1}^{\mars}_\text{pro}  - \text{F1}^{\mars}_\text{anti} \right| + \left| \text{F1}^{\female}_\text{pro}  - \text{F1}^{\female}_\text{anti} \right| \right)
\end{align*}
\noindent where $\text{F1}^{\mars}_\text{pro}$ denotes the F1 score on the pro-stereotypical dataset whilst considering the male pronoun as the \emph{true} label. To be completely gender neutral, we average the absolute values since we are only interested in the extent of gender bias rather than its direction. 
\subsection{Online Skewness Mitigation}\label{subsec:naive}
As we will show in \cref{subsec:winobias-performance}, most current TBTL models models are inherently skewed towards predicting male pronouns.
Inspired by \citet{kurita2019measuring}, we propose a simple approach to reducing this skew.
We normalise the probability of a masked pronoun being assigned a particular gender in a certain occupational context by dividing through with the prior probability of choosing that pronoun in a sentence with the same structure but without any occupational context.
We illustrate this method with the sentence ``The physician hired the secretary because he was overwhelmed with clients''. This method starts by calculating the probabilities of ``he'' and ``she'' in the standard way, as described in \cref{subsec:method1}.
Next, we mask the professions, leading to ``[MASK] hired [MASK] because [MASK] was overwhelmed with clients'' and calculate the probability of the third masked word being ``he'' and ``she'' in this context. 
Finally, we normalise by dividing the probabilities found using the standard method, with the probabilities found using the masked-professions context.
This method assumes language models can resolve the pronoun when both professions are masked.

Models mitigating skew using this approach are given the suffix \emph{-O} in the remainder of this paper.

\subsection{Bias Removal via Data Augmentation} \label{sec:method_data_aug}

We aim to replicate the Data Augmentation method proposed in \citet{zhao2019gender} for mitigating gender bias in ELMo.
The goal of this approach is to use an augmented dataset to fine-tune the pre-trained language model to the GPR task.
In particular, this augmented dataset is designed with the intention of neutralising the gender bias already present in a model such as BERT, whilst simultaneously avoiding the corruption of its understanding of natural language. 
As in \citet{zhao2019gender}, a target GPR task was constructed by first selecting sentences from OntoNotes 5.0 containing gendered pronouns and masking them accordingly; the BERT masked language model will be trained to predict the masked pronoun.
Secondly, we anonymise the data by replacing all gendered names with identity tokens such as [E1] and [E2].
Each training example is then \emph{augmented} by replacing all possessive and personal pronouns with those of the opposite gender. Additionally we apply a mapping of explicitly gendered words (such as ``Man''$\xrightarrow{}$``Woman'' and vice-versa) to ensure that the text remains linguistically coherent in the context of reversed genders.\footnote{Mapping sourced from~\citep{zhao2018gender}. See the \href{https://github.com/uclanlp/gn_glove/tree/master/wordlist}{Glove} and \href{https://github.com/uclanlp/corefBias/tree/master/WinoBias/wino}{WinoBias} GitHub pages.}
Following this approach, the sentence ``\emph{The King was pleased that his Lords had vanquished their enemies}'' would be augmented to ``\emph{The Queen was pleased that her Ladies had vanquished their enemies}''.

To examine the effects of data augmentation, we then fine-tune two BERT models. The first was fine-tuned on the un-augmented OntoNotes training examples, whilst the second was fine-tuned on the augmented OntoNotes examples (containing the duplicated and gender-switched examples also). Hereafter we shall refer to these models as BERT-U and BERT-A respectively. In both cases, a hyperparameter search over the epochs and learning rate was conducted.\footnote{Specific settings detailed in~\cref{suppdata}.} The best performing un-augmented/augmented models were tested using the WinoBias data as described in \cref{subsec:method1}. 

\section{Results and Discussion}\label{Sec:results}
\subsection{Baseline: Alice and Bob} \label{subsec:alicebob}

\begin{table}[t]
\caption{F1 (in \%) performance of different models on WinoBias dataset, where professions are replaced by gendered names and a pronoun is correct if it refers to the correct name. The insertion of gendered names implies that there is now a correct pronoun, in contrast to the original WinoBias data set where there is merely a stereotypical pronoun.
The suffixes O, U and A refer to Online, Un-augmented and Augmented mitigation approaches respectively.}
\label{tab:alicebob}
\begin{center}
\resizebox{\columnwidth}{!}{%
\begin{tabular}{lrrrr}

\toprule
\multirow{2}{*}{\textbf{Model}} & \multicolumn{2}{c}{\textbf{T1}} & \multicolumn{2}{c}{\textbf{T2}} \\\cline{2-5}
                      & \textbf{F1}$^{\male}$      &\textbf{F1}$^{\female}$     & \textbf{F1}$^{\male}$      &\textbf{F1}$^{\female}$ \\ \toprule 
 RoBERTa &       64.2 &       72.2 &       92.7 &       93.0  \\
   RoBERTa-O &       50.2 &       71.7 &       89.8 &       88.7  \\ \midrule
     RoBERTa-large &       78.2 &       79.5 &       94.9 &       94.9   \\
    RoBERTa-large-O &       78.2 &       79.5 &     89.5 &       87.3   \\ \midrule
           ALBERT &       39.2 &       68.6  & 58.7 &       75.3  \\
          ALBERT-O &        6.2 &       67.6 &        62.9 &       24.1    \\\midrule
        ALBERT-large &       61.0 &       71.2 &      19.2 &       68.6  \\
       ALBERT-large-O &       60.6 &       68.9 &      31.2 &       70.1   \\\midrule
       ALBERT-xlarge &       64.1 &       75.1 &    23.3 &       69.4   \\
     ALBERT-xlarge-O &       69.2 &       74.2 &    59.4 &       76.6  \\\midrule
      ALBERT-xxlarge &       64.2 &       76.9 &     95.0 &       95.3    \\
   ALBERT-xxlarge-O &       78.2 &       80.5 &     89.0 &       90.5   \\\midrule
                 BERT &       58.8 &       62.4 &   95.3 &       95.5    \\
              BERT-O &       59.0 &       64.7 &       95.1 &       95.1   \\\midrule
          BERT-large &       72.6 &       74.9 &        95.3 &       95.5    \\
       BERT-large-O &       69.8 &       75.1 &        95.6 &       95.7    \\\midrule
      XLM-RoBERTa &       29.7 &       69.0 &         64.5 &       76.7    \\
    XLM-RoBERTa-O &       52.1 &       65.1 &         64.4 &       37.9    \\\midrule
   XLM-RoBERTa-large &       62.8 &       76.4 &     21.7 &       69.2   \\
XLM-RoBERTa-large-O &       64.9 &       77.0 &       80.1 &       84.9   \\\midrule
         DistilBERT &       41.2 &       66.4 &      81.0 &       79.0   \\
    DistilBERT-O &       50.2 &       66.1 &     81.0 &       78.5   \\ \midrule
BERT-U & 76.6 & 65.8 & 90.1 & 88.6     \\ 
BERT-UO & 78.8 & 77.3 & 93.1 & 92.3     \\ \midrule 
BERT-A & 75.6 & 67.4 &  91.8 & 90.0      \\ 
BERT-AO & 75.7 & 67.5 & 74.9 & 54.2     \\ \bottomrule 
\end{tabular}
}
\end{center}
\end{table}

There is a risk that removing bias deteriorates the predictive power of the model. We measure a baseline performance on a GPR task to test how well the model is able to actually resolve the pronoun to the correct entity. To assess this we modify the WinoBias data set by replacing the professions with unambiguously gendered names, \textit{Alice} and \textit{Bob}. \cref{tab:alicebob} illustrates that we can achieve high F1 scores on this modified WinoBias dataset, validating the use of masked language models for GPR tasks. However, we note that ALBERT and XLM-RoBERTa perform particularly poorly on both T1 and T2 tasks.

The Online Skewness Mitigation described in \cref{subsec:naive} demonstrates no discernable pattern on the F1 scores. This suggests that it does not negatively effect GPR performance. Neither is there a definite pattern in the skew, indicating that skew is not necessarily reduced in the presence of unambiguously gendered entities.

The decreased performance of BERT-A/U (as described in \cref{sec:method_data_aug}) relative to the out-of-the-box BERT model may be caused by the anonymisation of the OntoNotes data the models were fine-tuned on, making them less receptive to performing GPR with common names.

The F1 scores in \cref{tab:alicebob} demonstrate that  GPR in T1 is significantly more challenging than in T2.  Figures~\ref{fig:winobias1} and~\ref{fig:winobias2} show that pronoun resolution in T2 is always with respect to \emph{[entity2]} whilst in T1, the pronoun resolution can be with respect to either \emph{[entity1]} or \emph{[entity2]}. In T2 it is clearer to contextual models which entity is the object of the sentence. Thus, we can use a model's gendered pronoun predictions on T2 sentences to expose any internal bias it may have toward \emph{[entity2]}. In T1, the lack of syntactic cues make it unclear which entity is the sentence's object; as such, we may be unable to isolate the model bias corresponding to each specific entity.\footnote{This also explains the larger skew on T1 compared to T2: the lower GPR performance on T1 leads the model to guess more and as we will see in \cref{subsec:winobias-performance}, prefers to guess male.} For these reasons, we argue that T2 is better at revealing the biases encoded in these models. \emph{Hence we will use T2 from this point onward.}

To expose bias, we require models with reliable coreference resolution performance on T2, demonstrated by consistently high F1 scores. In subsequent sections, we investigate all BERT, RoBERTa, and DistilBERT models as well as ALBERT-xxlarge and XLM-RoBERTa-large-O, all of which have F1 scores on T2 greater than our arbitrarily defined threshold of 75\%. All of the other models in \cref{tab:alicebob} fall below this threshold and so are not considered further in this paper.

\subsection{WinoBias Performance}\label{subsec:winobias-performance}

\begin{table}[t]
\caption{F1 results (in \%) from Test Set 2. The suffixes O, U and A refer to Online, Unaugmented and Augmented bias mitigation approaches respectively.}
\label{tab:test_F1}
\resizebox{0.5\textwidth}{!}{%
\begin{tabular}{lccccccc}
\toprule
          &  \multicolumn{6}{c}{\textbf{T2}}                                                                                                          \\ \cline{2-7} 
                        & \multicolumn{2}{c}{\textbf{F1}$^{\male}$} & \multicolumn{2}{c}{\textbf{F1}$^{\female}$} & \multicolumn{2}{c}{\textbf{Bias}}                          \\ \cline{2-7} 
\multirow{-3}{*}{\textbf{Embedding}} &                        Pro               & Anti             & Pro               & Anti             & Stereo                         & Skew                       \\
\toprule
     RoBERTa &       62.9 &        27.0 &       69.0 &        39.3 &        32.8 &       9.2 \\ 
   RoBERTa-O &       68.0 &        60.2 &       26.5 &         8.5 &        12.9 &      46.6 \\ \midrule
    RoBERTa-large &       67.0 &        52.4 &       45.0 &        21.5 &        26.4 &      24.0 \\ 
  RoBERTa-large-O &       66.0 &        65.0 &       11.2 &         7.5 &         2.3 &      56.1 \\ \midrule
   ALBERT-xxlarge &       71.4 &        46.9 &       54.8 &        16.4 &        31.4 &      23.5 \\ 
 ALBERT-xxlarge-O &       69.7 &        49.2 &       51.5 &        18.1 &        27.0 &      24.6 \\ \midrule
       DistilBERT &       64.9 &        67.2 &        4.8 &         5.0 &         1.3 &      61.2 \\
    DistilBERT-O &      64.5 &        65.8 &       10.0 &        12.8 &         2.1 &      53.8 \\ \midrule
        BERT &     69.3 &        58.0 &       31.4 &         8.2 &        17.3 &      43.8 \\
      BERT-O &       68.4 &        58.1 &       32.9 &        10.5 &        16.4 &      41.6 \\ \midrule
       BERT-large &        70.0 &        57.9 &       33.9 &         2.8 &        21.6 &      45.6 \\
     BERT-large-O &        69.9 &        57.9 &       32.5 &         5.0 &        19.7 &      45.1 \\ \midrule
     XLM-RoBERTa-large-O &       68.0 &        56.0 &       38.2 &        15.7 &        17.2 &      35.1 \\ \midrule
     BERT-U       & 67.8              & 63.4             & 14.1              & 2.8              & 57.2 & 7.9  \\ 
BERT-UO           & 67.3              & 60.4             & 26.6              & 11.6             & 44.8 & 11.0 \\ \midrule
BERT-A            & 64.7              & 64.5             & 14.8              & 14.8             & 49.8 & 0.1 \\ 
BERT-AO           & 65.0              & 63.6             & 17.9              & 15.3             & 47.7 & 2.0 \\
\bottomrule
\end{tabular}
}
\end{table}

In \cref{tab:test_F1} we present the F1 scores achieved on WinoBias T2 by different models. We note that the pro-stereotypical F1 scores are lower than the gendered names baseline of \cref{tab:alicebob}. This is to be expected, since the Alice and Bob system discussed in \cref{subsec:alicebob} can be understood as being unambiguous and completely biased. Consequently, whereas \cref{tab:alicebob} shows just GPR performance and general skew bias, \cref{tab:test_F1} quantifies GPR performance, skew and stereotype.

Note that all models show significant male skew, except for RoBERTa which demonstrates higher F1$^{\female}$ than F1$^{\male}$ scores on both pro- and anti-stereotypical examples. Indeed for all other models there is a noticeable increase in male skew compared to the Alice \& Bob results in \cref{tab:alicebob}. The only experimental difference is the use of occupations rather than names, demonstrating that it is specifically the professions that push the model to predicting male pronouns.

Focusing on the out-of-the-box models, we rank them by their gender skew from best to worst as RoBERTa, ALBERT-xxlarge, RoBERTa-large, BERT, BERT-large, and DistilBERT. 

We note that RoBERTa has the least skew bias, with a $\mu_{\text{skew}}$ value of 9.2\% for T2.
\citet{roberta} report that BERT was ``significantly undertrained'', and aimed to address this by training RoBERTa for longer, with bigger batches and sequences, additional data, and dynamic adjustments to the masking pattern.
These amendments in RoBERTa appear to have reduced the skew bias in the model, suggesting that a model's training procedure can have a considerable impact on its skew. We also observed that within the BERT and RoBERTa families, larger models tend to show more skew than their smaller counterparts.

The high skew of DistilBERT might be due to its student-teacher training~\citep{hinton2015distilling}. This lends itself to a overly simplistic understanding of male and female roles within society.  Understanding the subtleties and nuances of gender roles requires models with high representation capacity and training DistilBERT to mimic BERT's output renders it incapable of making such distinctions.

\begin{figure}[t]
\begin{center}
\includegraphics[width=0.45\textwidth]{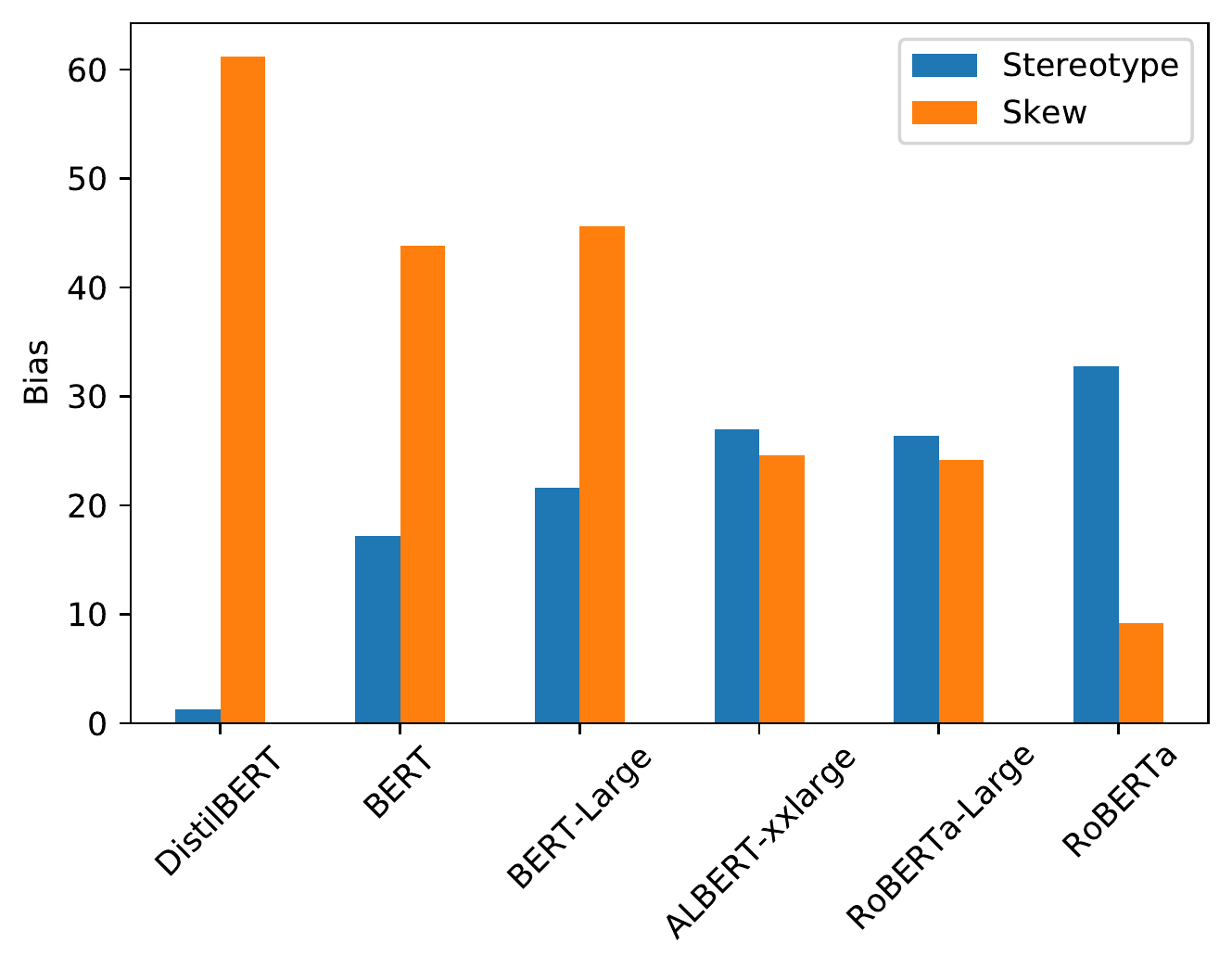}
\caption{Bar chart showing the different bias measurements for the out-of-the-box models investigated. This suggests an inherent trade-off between skew and stereotype in language models. }
\label{fig:biasbargraph}
\end{center}
\end{figure}

The ranking of gender stereotype from best to worst is DistilBERT, BERT, BERT-large, RoBERTa-large, ALBERT-xxlarge, and RoBERTa.

Note that this order is approximately the opposite to skew, as illustrated in \cref{fig:biasbargraph}. There appears to be a potential trade off between the skew and stereotype in out-of-the-box language models, with RoBERTa-large best balancing the two biases. This trend appears to carry forward to the fine-tuned models, with BERT-A and BERT-U showing high stereotype but very low skew. 

\subsection{Mitigated Systems}
\subsubsection{Online Skewness Mitigation}

Comparing how bias values change in \cref{tab:test_F1} when going from all models to their normalised - O version, Online Skewness Mitigation successfully reduces stereotype for 6/8 models, though interestingly RoBERTa responded by going from a female stereotype to a significant male stereotype. At the same time, we observe that this reduction in model stereotype actually comes at the expense of model skew. Note that this effect is the opposite to what it was designed for. For this reason we do not explore it further.

\subsubsection{Data Augmentation}
\label{sec:results-dataaug}

BERT-U and BERT-A test accuracies on the OntoNotes data were 64\% and 58\% respectively. This 6\% loss in accuracy should be kept in mind when assessing gender bias. We see BERT-U/A have similar F1 values in \cref{tab:alicebob} indicating that any difference in bias is not just noise coming from them having radically different abilities to perform the task.

\cref{tab:test_F1} shows BERT-A has a 7.4\% lower stereotype and a 7.8\% lower skew than BERT-U, echoing the findings of \citet{zhao2019gender} which demonstrate reductions of bias in ELMo through fine-tuning with data augmentation. The skew of BERT-A is reduced to 0.1\%; much lower than standard BERT. This elimination of skew is unsurprising since augmentation ensures professional entities assume male and female roles equally in the fine-tuning data. Despite this, the stereotype of BERT-A is greater than standard BERT, suggesting a degree of non-skew bias has been introduced by fine-tuning. An analysis of the individual test examples in WinoBias T2 may reveal why this is the case. Across the classifications of sentences we see in the data, pronoun resolution appears to be influenced by the implied characteristics of the individual whose pronouns we seek to resolve, rather than the professions themselves.

\subsection{Impact of Sentence Context on Pronoun Resolution}
\label{sec:results-competency}
We consider an example from WinoBias T2 with the \emph{[entity2]} profession being ``guard''. The only example where BERT-A assigned a female pronoun to the profession ``guard'' was a sentence where the guard was associated with low competency: ``\emph{The clerk works harder than the guard and gets more appreciation than [her]}''. To investigate this more quantitatively, we measure the competency of the entity whose pronoun is being resolved. Our aim is to identify whether fine-tuning results in a gender shift across competent and incompetent examples which might suggest a competency-based gender assignment.

To achieve this, four of the authors labelled the WinoBias examples according to whether they thought the subject of the sentence demonstrated competent or skilful behaviour (e.g. getting a promotion, being thanked for a job well done) or not (e.g. getting sacked from their job). The sentences were classified as Incompetent, Neutral (no competency suggested in the sentence), or Competent. All voters classified sentences independently, with a Fleiss' $\kappa$ score of 0.42. The class assigned to each sentence was then determined by a majority vote. Sentences that resulted in a tie were discarded.\footnote{Our competency dataset is available at \href{https://github.com/12kleingordon34/NLP_masters_project}{GitHub}.} To isolate our investigation of subject competency from professional stereotype, all professions in the WinoBias dataset were replaced with the gender agnostic term ``person''.

\cref{tab:competency} reports the proportion of examples in each competency class that were assigned a female pronoun across BERT, BERT-U, and BERT-A. The proportions of female pronoun assignments show that BERT-A allocates a more balanced ratio of gendered pronouns to Incompetent examples compared to BERT and BERT-U. Apart from the Competent class (which shows no major change across all three models), BERT-A reduces the gender imbalance of pronouns in Neutral and Incompetent examples.

It is challenging to exactly determine the cause of these observations, but it certainly appears that fine-tuning BERT models has an effect on the gender ratios in each competency class. It is notable that de-biasing BERT reduced the gender imbalance of Incompetent examples by a large margin. These findings merit further investigation.

We believe that WinoBias and other related benchmarks do not sufficiently probe professional gender bias, as pronoun resolution may be obfuscated by cross-correlations from other manifestations of gender prejudice. One example of a bias other than profession and competency could be personality bias, where women may be more closely associated with passive and caring traits whilst men may be more aggressive and disagreeable. We encourage the development of a dataset that isolates these different gender biases, allowing us to probe them without interference from one another.

\begin{table}[t]
\begin{tabular}{lccccc}
\toprule
{\bf Competency} & {\bf BERT}  & {\bf BERT-U} & {\bf BERT-A} \\ \midrule
 Incompetent         & 0.156 & 0.062  & 0.281  \\ 
                       Neutral          & 0.117 & 0.168  & 0.280  \\
                       Competent           & 0.160 & 0.140   & 0.140  \\ \bottomrule
\end{tabular}
\caption{Proportion of female pronouns assigned for each competency category for the WinoBias T2 sentences. All professional entities in the examples were replaced with the gender neutral term ``person'' to isolate the impact of competency from professional stereotype.}
\label{tab:competency}
\end{table}

\section{Conclusion and Future Work}\label{Sec:conclusion}
Quantifying gender bias in coreference resolution is challenging, since co-referencing performance and bias manifestation are closely linked. We have proposed \emph{skew} and \emph{stereotype} as new measures of gender bias, allowing us to better probe model prejudice. 

We have shown that there is an approximate trade-off between the skew and stereotype of out-of-the-box models. DistilBERT and BERT models have high skew and low stereotype whilst RoBERTa and ALBERT-xxlarge have reduced skew at the cost of higher stereotype.

Two methods have been proposed to mitigate bias: Online Skewness Mitigation and Data Augmentation. The online approach has been shown to be effective at mitigating stereotype at the expense of skew, demonstrating the opposite effect to what it was designed for. We took the Data Augmentation method proposed by \citet{zhao2019gender} for debiasing ELMo and extended it to BERT, demonstrating that it reduces both forms of gender bias compared to unaugmented fine-tuned models. However, the reduction of explicit professional gender skew and stereotype reveal the model's underlying bias towards gender competency. We successfully expose these using WinoBias GPR sentence probes labelled for competency. 

Since contextual language models consider the full sentence contents when assigning a pronoun, we believe that the WinoBias data used in this paper does not purely measure professional biases. A second popular dataset taken from the SuperGLUE benchmarks, \emph{Winogender} \citep{rudinger2018gender}, is increasingly used for evaluating a model's gender bias. However, its limited size relative to WinoBias makes it less robust and hence it was not used in this paper. 

We observed that language models may also consider other stereotyped gender characteristics in the sentence when classifying pronouns. Given the above considerations, we believe that a more comprehensive set of gender bias benchmarks should be developed which can better isolate specific biases within models. 

\citet{kiritchenko-mohammad-2018-examining} have shown that both race and gender bias are prevalent in a large proportion of state-of-the-art language models. Recently, a number of other datasets have appeared for detecting these and other kinds of bias such as age and religion \citep{nadeem2020stereoset, nangia-etal-2020-crows}. It would be interesting to see if competency bias obscures analyses on these datasets similarly.

Future research is recommended on how data augmentation affects other models from the BERT family. Additionally, it will be valuable to explore whether Data Augmentation could be applied to larger corpora to train a new contextual model from scratch.

Lastly, in contrast to static embeddings, it is notoriously hard, if not impossible, to define bias in contextual embeddings~\cite{Caliskan}. It is likely that without extensive research and transparent communication, the field of NLP will be further scrutinised as more applications are found to exhibit undesired biases. Discussions, both within and outside the community, are required to determine what separates bias from semantic assumptions, allowing bias disclaimers and guidelines to be provided to downstream developers.

\subsubsection*{Acknowledgements}
We thank the anonymous reviewers for the insightful comments, and Hadas Orgad for helping us correcting a mistake in a previous version of this paper.
This research was supported by the European Union's Horizon 2020 research and innovation programme under grant agreement no. 875160.

\bibliography{bibliography.bib}

\begin{thebibliography}{31}
\expandafter\ifx\csname natexlab\endcsname\relax\def\natexlab#1{#1}\fi

\bibitem[{Bolukbasi et~al.(2016)Bolukbasi, Chang, Zou, Saligrama, and
  Kalai}]{bolukbasi2016man}
Tolga Bolukbasi, Kai-Wei Chang, James~Y Zou, Venkatesh Saligrama, and Adam~T
  Kalai. 2016.
\newblock \href
  {https://doi.org/https://dl.acm.org/doi/10.5555/3157382.3157584} {Man is to
  computer programmer as woman is to homemaker? debiasing word embeddings}.
\newblock \emph{Advances in Neural Information Processing Systems}, pages
  4349--4357.

\bibitem[{Caliskan et~al.(2017)Caliskan, Bryson, and Narayanan}]{Caliskan}
Aylin Caliskan, Joanna~J. Bryson, and Arvind Narayanan. 2017.
\newblock \href {https://doi.org/10.1126/science.aal4230} {Semantics derived
  automatically from language corpora contain human-like biases}.
\newblock \emph{Science}, 356:183–186.

\bibitem[{Chelba et~al.(2013)Chelba, Mikolov, Schuster, Ge, Brants, and
  Koehn}]{1billionwords}
Ciprian Chelba, Tomas Mikolov, Mike Schuster, Qi~Ge, Thorsten Brants, and
  Phillipp Koehn. 2013.
\newblock \href {http://arxiv.org/abs/1312.3005} {One billion word benchmark
  for measuring progress in statistical language modeling}.
\newblock \emph{CoRR}, abs/1312.3005.

\bibitem[{Dastin(2018)}]{dastin}
Jeffrey Dastin. 2018.
\newblock \href
  {https://www.reuters.com/article/us-amazon-com-jobs-automation-insight/amazon-scraps-secret-ai-recruiting-tool-that-showed-bias-against-women-idUSKCN1MK08G}
  {Amazon scraps secret {AI} recruiting tool that showed bias against women}.
\newblock \emph{Reuters}.

\bibitem[{Devlin et~al.(2018)Devlin, Chang, Lee, and Toutanova}]{bert}
Jacob Devlin, Ming{-}Wei Chang, Kenton Lee, and Kristina Toutanova. 2018.
\newblock \href {http://arxiv.org/abs/1810.04805} {{BERT:} pre-training of deep
  bidirectional transformers for language understanding}.
\newblock \emph{CoRR}, abs/1810.04805.

\bibitem[{Gonen and Goldberg(2019)}]{gonen2019lipstick}
Hila Gonen and Yoav Goldberg. 2019.
\newblock \href {https://doi.org/https://www.aclweb.org/anthology/N19-1061/}
  {Lipstick on a pig: Debiasing methods cover up systematic gender biases in
  word embeddings but do not remove them}.
\newblock \emph{Association for Computational Linguistics}, pages 609--614.

\bibitem[{Hinton et~al.(2015)Hinton, Vinyals, and Dean}]{hinton2015distilling}
Geoffrey Hinton, Oriol Vinyals, and Jeffrey Dean. 2015.
\newblock \href {http://arxiv.org/abs/1503.02531} {Distilling the knowledge in
  a neural network}.
\newblock \emph{Advances in Neural Information Processing Systems}.

\bibitem[{Kiritchenko and Mohammad(2018)}]{kiritchenko-mohammad-2018-examining}
Svetlana Kiritchenko and Saif Mohammad. 2018.
\newblock \href {https://doi.org/10.18653/v1/S18-2005} {Examining gender and
  race bias in two hundred sentiment analysis systems}.
\newblock \emph{Association for Computational Linguistics}, pages 43--53.

\bibitem[{Kurita et~al.(2019)Kurita, Vyas, Pareek, Black, and
  Tsvetkov}]{kurita2019measuring}
Keita Kurita, Nidhi Vyas, Ayush Pareek, Alan~W Black, and Yulia Tsvetkov. 2019.
\newblock \href {https://doi.org/10.18653/v1/W19-3823} {Measuring bias in
  contextualized word representations}.
\newblock \emph{Association for Computational Linguistics}, pages 166--172.

\bibitem[{Lan et~al.(2020)Lan, Chen, Goodman, Gimpel, Sharma, and
  Soricut}]{DBLP:conf/iclr/LanCGGSS20}
Zhenzhong Lan, Mingda Chen, Sebastian Goodman, Kevin Gimpel, Piyush Sharma, and
  Radu Soricut. 2020.
\newblock \href
  {https://doi.org/https://iclr.cc/virtual_2020/poster_H1eA7AEtvS.html}
  {{ALBERT:} {A} lite {BERT} for self-supervised learning of language
  representations}.
\newblock \emph{ICLR}.

\bibitem[{Lee et~al.(2017)Lee, He, Lewis, and Zettlemoyer}]{leegloveresult}
Kenton Lee, Luheng He, Mike Lewis, and Luke Zettlemoyer. 2017.
\newblock \href {http://arxiv.org/abs/1707.07045} {End-to-end neural
  coreference resolution}.
\newblock \emph{CoRR}, abs/1707.07045.

\bibitem[{Lee et~al.(2018)Lee, He, and Zettlemoyer}]{kentonleecoref}
Kenton Lee, Luheng He, and Luke Zettlemoyer. 2018.
\newblock \href {http://arxiv.org/abs/1804.05392} {Higher-order coreference
  resolution with coarse-to-fine inference}.
\newblock \emph{CoRR}, abs/1804.05392.

\bibitem[{Levesque et~al.(2012)Levesque, Davis, and
  Morgenstern}]{levesque_winograd_2012}
Hector~J. Levesque, Ernest Davis, and Leora Morgenstern. 2012.
\newblock \href {https://cs.nyu.edu/faculty/davise/papers/WSKR2012.pdf} {The
  {Winograd} {Schema} {Challenge}}.
\newblock In \emph{Proceedings of the {Thirteenth} {International} {Conference}
  on {Principles} of {Knowledge} {Representation} and {Reasoning}}, {KR}'12,
  pages 552--561. AAAI Press, Rome, Italy.

\bibitem[{Liu et~al.(2019)Liu, Ott, Goyal, Du, Joshi, Chen, Levy, Lewis,
  Zettlemoyer, and Stoyanov}]{roberta}
Yinhan Liu, Myle Ott, Naman Goyal, Jingfei Du, Mandar Joshi, Danqi Chen, Omer
  Levy, Mike Lewis, Luke Zettlemoyer, and Veselin Stoyanov. 2019.
\newblock \href {http://arxiv.org/abs/1907.11692} {Roberta: {A} robustly
  optimized {BERT} pretraining approach}.
\newblock \emph{CoRR}, abs/1907.11692.

\bibitem[{Lu et~al.(2018)Lu, Mardziel, Wu, Amancharla, and
  Datta}]{Lu_contextual_bias}
Kaiji Lu, Piotr Mardziel, Fangjing Wu, Preetam Amancharla, and Anupam Datta.
  2018.
\newblock \href {http://arxiv.org/abs/1807.11714} {Gender bias in neural
  natural language processing}.
\newblock \emph{CoRR}, abs/1807.11714.

\bibitem[{May et~al.(2019)May, Wang, Bordia, Bowman, and
  Rudinger}]{may2019measuring}
Chandler May, Alex Wang, Shikha Bordia, Samuel~R Bowman, and Rachel Rudinger.
  2019.
\newblock \href {https://doi.org/10.18653/v1/N19-1063} {On measuring social
  biases in sentence encoders}.
\newblock \emph{Association for Computational Linguistics}, pages 622--628.

\bibitem[{Nadeem et~al.(2020)Nadeem, Bethke, and Reddy}]{nadeem2020stereoset}
Moin Nadeem, Anna Bethke, and Siva Reddy. 2020.
\newblock \href {https://doi.org/https://arxiv.org/abs/2004.09456} {Stereoset:
  Measuring stereotypical bias in pretrained language models}.
\newblock \emph{arXiv preprint: 2004.09456}.

\bibitem[{Nangia et~al.(2020)Nangia, Vania, Bhalerao, and
  Bowman}]{nangia-etal-2020-crows}
Nikita Nangia, Clara Vania, Rasika Bhalerao, and Samuel~R. Bowman. 2020.
\newblock \href {https://doi.org/10.18653/v1/2020.emnlp-main.154}
  {{C}row{S}-pairs: A challenge dataset for measuring social biases in masked
  language models}.
\newblock \emph{Empirical Methods in Natural Language Processing}, pages
  1953--1967.

\bibitem[{Rudinger et~al.(2018)Rudinger, Naradowsky, Leonard, and
  Van~Durme}]{rudinger2018gender}
Rachel Rudinger, Jason Naradowsky, Brian Leonard, and Benjamin Van~Durme. 2018.
\newblock \href {https://doi.org/10.18653/v1/N18-2002} {Gender bias in
  coreference resolution}.
\newblock \emph{Association for Computational Linguistics}, pages 8--14.

\bibitem[{Shoeybi et~al.(2019)Shoeybi, Patwary, Puri, LeGresley, Casper, and
  Catanzaro}]{megatron}
Mohammad Shoeybi, Mostofa Patwary, Raul Puri, Patrick LeGresley, Jared Casper,
  and Bryan Catanzaro. 2019.
\newblock \href {http://arxiv.org/abs/1909.08053} {Megatron-lm: Training
  multi-billion parameter language models using model parallelism}.
\newblock \emph{CoRR}, abs/1909.08053.

\bibitem[{Sun et~al.(2019)Sun, Gaut, Tang, Huang, ElSherief, Zhao, Mirza,
  Belding, Chang, and Wang}]{sunlitreview}
Tony Sun, Andrew Gaut, Shirlyn Tang, Yuxin Huang, Mai ElSherief, Jieyu Zhao,
  Diba Mirza, Elizabeth Belding, Kai-Wei Chang, and William~Yang Wang. 2019.
\newblock \href {https://doi.org/10.18653/v1/P19-1159} {Mitigating gender bias
  in natural language processing: Literature review}.
\newblock \emph{Association for Computational Linguistics}, pages 1630--1640.

\bibitem[{Tan and Celis(2019)}]{tan2019assessing}
Yi~Chern Tan and L~Elisa Celis. 2019.
\newblock \href {https://doi.org/https://arxiv.org/abs/1911.01485} {Assessing
  social and intersectional biases in contextualized word representations}.
\newblock \emph{Advances in Neural Information Processing Systems}, pages
  13230--13241.

\bibitem[{Tashea(2017)}]{tashea_17AD}
Jason Tashea. 2017.
\newblock \href
  {https://www.wired.com/2017/04/courts-using-ai-sentence-criminals-must-stop-now/}
  {Courts are using ai to sentence criminals. that must stop now}.
\newblock \emph{WIRED}.

\bibitem[{{U.S. Bureau of Labor Statistics}(2017)}]{laborstats}
{U.S. Bureau of Labor Statistics}. 2017.
\newblock Labor force statistics from the current population survey, 2017.
\newblock \url{https://www.bls.gov/cps/cpsaat11.htm}.

\bibitem[{Wagner et~al.(2016)Wagner, Graells-Garrido, Garcia, and
  Menczer}]{Wagner_2016}
Claudia Wagner, Eduardo Graells-Garrido, David Garcia, and Filippo Menczer.
  2016.
\newblock \href {https://doi.org/10.1140/epjds/s13688-016-0066-4} {Women
  through the glass ceiling: gender asymmetries in wikipedia}.
\newblock \emph{EPJ Data Science}, 5(1).

\bibitem[{Webster et~al.(2018)Webster, Recasens, Axelrod, and
  Baldridge}]{webster2018mind}
Kellie Webster, Marta Recasens, Vera Axelrod, and Jason Baldridge. 2018.
\newblock \href {https://doi.org/https://www.aclweb.org/anthology/Q18-1042/}
  {Mind the gap: A balanced corpus of gendered ambiguous pronouns}.
\newblock \emph{Association for Computational Linguistics}, 6:605–618.

\bibitem[{Weischedel et~al.(2011)Weischedel, Hovy, Marcus, Palmer, Belvin,
  Pradhan, Ramshaw, and Xue}]{ontonotes}
Ralph Weischedel, Eduard Hovy, Mitchell Marcus, Martha Palmer, Robert Belvin,
  Sameer Pradhan, Lance Ramshaw, and Nianwen Xue. 2011.
\newblock \emph{OntoNotes: A Large Training Corpus for Enhanced Processing}.

\bibitem[{Zhao et~al.(2019)Zhao, Wang, Yatskar, Cotterell, Ordonez, and
  Chang}]{zhao2019gender}
Jieyu Zhao, Tianlu Wang, Mark Yatskar, Ryan Cotterell, Vicente Ordonez, and
  Kai-Wei Chang. 2019.
\newblock \href {https://www.aclweb.org/anthology/N19-1064/} {Gender bias in
  contextualized word embeddings}.
\newblock \emph{Association for Computational Linguistics}, pages 629--634.

\bibitem[{Zhao et~al.(2018{\natexlab{a}})Zhao, Wang, Yatskar, Ordonez, and
  Chang}]{zhao2018gender}
Jieyu Zhao, Tianlu Wang, Mark Yatskar, Vicente Ordonez, and Kai-Wei Chang.
  2018{\natexlab{a}}.
\newblock \href {https://doi.org/10.18653/v1/N18-2003} {Gender bias in
  coreference resolution: Evaluation and debiasing methods}.
\newblock \emph{Association for Computational Linguistics}, pages 15--20.

\bibitem[{Zhao et~al.(2018{\natexlab{b}})Zhao, Zhou, Li, Wang, and
  Chang}]{zhao2018glove}
Jieyu Zhao, Yichao Zhou, Zeyu Li, Wei Wang, and Kai-Wei Chang.
  2018{\natexlab{b}}.
\newblock \href {https://doi.org/10.18653/v1/D18-1521} {Learning gender-neutral
  word embeddings}.
\newblock \emph{Association for Computational Linguistics}, pages 4847--4853.

\bibitem[{Zhu et~al.(2015)Zhu, Kiros, Zemel, Salakhutdinov, Urtasun, Torralba,
  and Fidler}]{bookcorpus}
Yukun Zhu, Ryan Kiros, Richard~S. Zemel, Ruslan Salakhutdinov, Raquel Urtasun,
  Antonio Torralba, and Sanja Fidler. 2015.
\newblock \href {http://arxiv.org/abs/1506.06724} {Aligning books and movies:
  Towards story-like visual explanations by watching movies and reading books}.
\newblock \emph{CoRR}, abs/1506.06724.

\end{thebibliography}
\bibliographystyle{acl_natbib}

\newpage
\appendix

\section{Fine-tuning Training Parameters}\label{suppdata}

For fine-tuning BERT on the OntoNotes data, the following settings were used. Standard hyperparameter choices of $\beta_1=0.9, \beta_2=0.999, \epsilon=10^{-8}$, and a dropout probability of 0.1 were chosen. Model training and validating with a 80/20 train/test split of the training data, across training epochs $\in \{1,\dots, 10\}$. The selected (epoch) model was that with the highest pronoun prediction accuracy on the validation set.

\end{document}